\documentclass[letterpaper]{article} 
\usepackage{aaai2026}  
\usepackage{times}  
\usepackage{helvet}  
\usepackage{courier}  
\usepackage[hyphens]{url}  
\usepackage{graphicx} 
\usepackage{natbib}  
\usepackage{caption} 
\usepackage{algorithm}
\usepackage{algorithmic}

\usepackage{newfloat}
\usepackage{listings}
\DeclareCaptionStyle{ruled}{labelfont=normalfont,labelsep=colon,strut=off} 
\floatstyle{ruled}
\newfloat{listing}{tb}{lst}{}
\floatname{listing}{Listing}
\usepackage{booktabs}       
\usepackage{bm}
\usepackage{amsmath}
\usepackage{amssymb}
\usepackage{cleveref}
\usepackage{multirow}
\usepackage{enumitem}
\usepackage{subcaption}
\usepackage{amsthm}
\newtheorem{corollary}{Corollary}
\usepackage{makecell}
\usepackage[misc]{ifsym}
\usepackage{adjustbox}

\newcommand{\ie}{\emph{i.e.},~}

\newcommand{\wrt}{\emph{w.r.t.}~}

\renewcommand{\paragraph}[1]{\medskip\noindent\textbf{#1.~}}

\newcommand{\zhongk}[1]{\left[#1\right]}

\newcommand{\calD}{\mathcal{D}}
\newcommand{\calE}{\mathcal{E}}

\newcommand{\calN}{\mathcal{N}}

\newcommand{\calS}{\mathcal{S}}

\newcommand{\bbE}{\mathbb{E}}

\newcommand{\bbR}{\mathbb{R}}

\newcommand{\modelname}{HUG}

\title{Heterogeneous Uncertainty-Guided Composed Image Retrieval with Fine-Grained Probabilistic Learning}
\author{
    Haomiao Tang\textsuperscript{\rm 1}, 
    Jinpeng Wang\textsuperscript{\rm 2}\thanks{Corresponding author.}, 
    Minyi Zhao\textsuperscript{\rm 3}, 
    Guanghao Meng\textsuperscript{\rm 1}, 
    Ruisheng Luo\textsuperscript{\rm 1}, \\
    Long Chen\textsuperscript{\rm 4}, 
    Shu-Tao Xia\textsuperscript{\rm 1}
}
\affiliations{
    \textsuperscript{\rm 1} Tsinghua Shenzhen International Graduate School, Tsinghua University\\
    \textsuperscript{\rm 2} Harbin Institute of Technology, Shenzhen \\
    \textsuperscript{\rm 3} Fudan University \\
    \textsuperscript{\rm 4} The Hong Kong University of Science and Technology \\
    thm23@mails.tsinghua.edu.cn, wjp20@mails.tsinghua.edu.cn
}

\begin{document}

\maketitle
\begin{abstract}
Composed Image Retrieval (CIR) enables image search by combining a reference image with modification text. 
Intrinsic noise in CIR triplets incurs intrinsic uncertainty and threatens model's robustness.
Probabilistic learning approaches have shown promise in addressing such issues; however, they fall short for CIR due to their instance-level holistic modeling and homogeneous treatments for queries and targets.
This paper introduces a \underline{\textbf{\textbf{H}}}eterogeneous \underline{\textbf{\textbf{U}}}ncertainty-\underline{\textbf{\textbf{G}}}uided (\textbf{\modelname{}}) paradigm to overcome these limitations. 
\modelname{} utilizes a \emph{fine-grained} probabilistic learning framework, where queries and targets are represented by Gaussian embeddings capturing detailed concepts and uncertainties. 
We customize \emph{heterogeneous} uncertainty estimations for multi-modal queries and uni-modal targets.
Given a query, we capture uncertainties not only regarding uni-modal content quality but also multi-modal coordination, followed by a \emph{provable} dynamic weighting mechanism to derive the comprehensive query uncertainty. 
We further design uncertainty-guided objectives, including query-target holistic contrast and fine-grained contrasts with comprehensive negative sampling strategies, which effectively enhance discriminative learning.
Experiments on benchmarks demonstrate \modelname{}'s effectiveness beyond state-of-the-art baselines, with faithful analysis justifying the technical contributions. 
\end{abstract}

\begin{links}
    \link{Code}{https://github.com/tanghme0w/AAAI26-HUG}
\end{links}
\section{Introduction}
\label{sec: introduction}

\begin{figure}[ht]
    \centering
    \includegraphics[width=\linewidth]{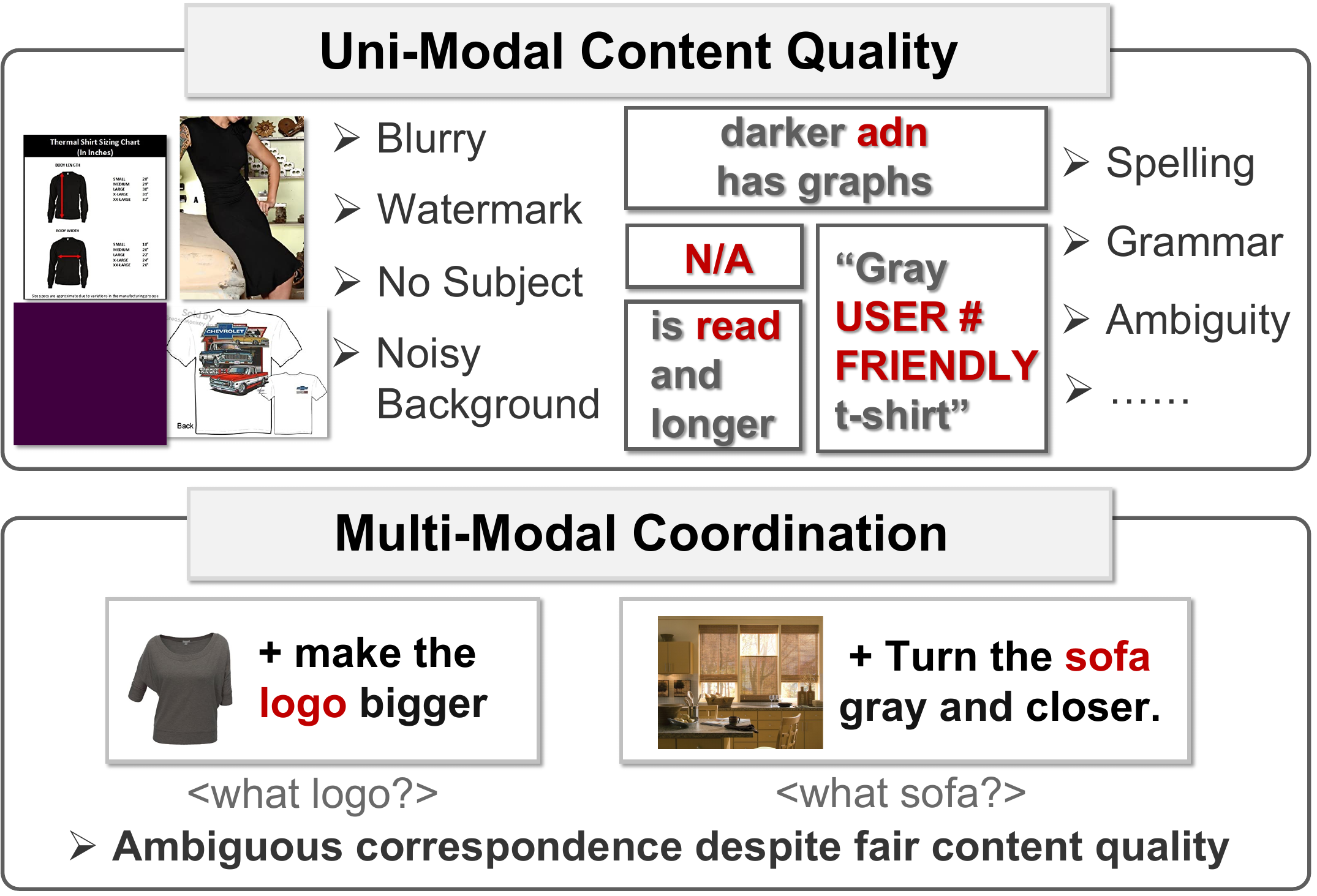}
    \caption{In Composed Image Retrieval, the uncertain multi-modal coordination between the reference image and modification text is also important in representation learning.}
    \label{fig:intro}
\end{figure}

Composed Image Retrieval (CIR) ~\cite{vo2019composing} is an emerging topic in multimedia retrieval that allows searching for images with multi-modal queries comprising reference images and modification texts. 
It allows users to articulate complex visual preferences that might be difficult to express through text or images alone, which facilitates personalization and is favorable in various e-commerce applications and social media~\cite{wu2021fashion}. 
Despite the practical value, CIR is more challenging than classic uni-modal \cite{Bowyer2000A2A, Wan2014DeepLF, Dubey2020ADS,autossvh2025} or cross-modal \cite{Lee2018StackedCA, Li2019VisualSR, hcq2022,hugspp2024,lase2026} retrieval tasks on learning robust representations. 
This inherently results in uncertainty that threatens the robustness of search models.

The uncertainty in CIR is heterogeneous. We can characterize it by two typical forms, as exemplified in \Cref{fig:intro}: 
\begin{enumerate}[label=(\textbf{\roman*}),leftmargin=1.8em]
    \item \textbf{\emph{Content Quality}}. Low-quality elements, such as blurry images or uninformative texts, are hard to avoid in CIR. 
    
    \item \textbf{\emph{Multi-Modal Coordination within Queries}}. 
    In CIR, the multi-modal nature of queries raises a particular coordination issue. Even if an image and its accompanying 
    text are considered high-quality individually, there may still be an ambiguous correspondence or mismatch. 
\end{enumerate}
Note that related works in multi-modal retrieval ~\cite{song2019polysemous, chun2021probabilistic, andrei2022probabilistic, chun2024improved, cope2025} have provided some inspiration by probabilistic embedding learning \cite{Abdar2020ARO, oh2019modeling}, which helps to identify and handle some of the above issues via uncertainty estimation.
However, existing solutions still exhibit two major drawbacks when applied to CIR. 
Firstly, they typically operate at an \emph{instance granularity}, failing to capture the complex and fine-grained user intents in CIR. 
Secondly, they apply \emph{homogeneous strategies} to both query and target sides, since both are uni-modal. 
This may not be the best practice in CIR because the multi-modal coordination uncertainty issue at the query side requires further remedy. 

In this paper, we propose a \underline{\textbf{\textbf{H}}}eterogeneous \underline{\textbf{\textbf{U}}}ncertainty-\underline{\textbf{\textbf{G}}}uided paradigm (\textbf{\modelname{}}) to comprehensively address these issues. 
\modelname{} is carefully developed with a \emph{fine-grained} probabilistic learning framework, representing each query and target image as a series of Gaussian embeddings. 
Each Gaussian aims to describe a fine-grained detail and capture a latent concept in the intricate matching space. 
The variance reflects the fine-grained uncertainty, allowing models to prioritize certain details while mitigating the adverse effects of fuzzy ones in the matching process. 
To better target CIR, we develop \emph{heterogeneous} uncertainty estimation for the uni-modal target and the multi-modal query: while the target side only needs to model the content quality uncertainty, the query side further considers the multi-modal coordination uncertainty between the reference image and the modification text. 
In particular, to obtain overall query uncertainty, we combine the text- and image-specific content quality uncertainties as well as the multi-modal coordination uncertainty through a provable dynamic weighting mechanism. 
Guided by the established estimations, we introduce uncertainty-aware contrastive loss to learn discriminative holistic matching between queries and targets. 
Moreover, we further design uncertainty-guided fine-grained contrast for each Gaussian embedding, incorporating \emph{component-}, \emph{instance-}, and \emph{modality-wise} negative sampling strategies to supplement robust learning signals. 

We conduct extensive experiments on standard CIR benchmarks, showing \modelname{}'s effectiveness against state-of-the-art baselines. 
Besides, we present a detailed model analysis, examining contributions of the key designs in \modelname{}, including fine-grained representation, heterogeneous uncertainty estimation, and uncertainty-guided objectives. 
Moreover, the quantitative study of the learned representations via \modelname{} reveals that each component of uncertainty can intuitively reflect image or text attributes, such as color, logo, or sleeve length, while the magnitude of uncertainty closely correlates with the ambiguity of these aspects. 
These intuitive findings highlight an intriguing interpretability in \modelname{}.

To summarize, we make the following contributions: 
\setlist{nolistsep}
\begin{itemize}[leftmargin=1em]
\item \textbf{\emph{Fine-grained probabilistic representation}}: We represent each query and target image as a series of Gaussian embeddings to better capture attribute-level details, where variances reflect fine-grained uncertainties and help prioritize important details during the matching process.

\item \textbf{\emph{Heterogeneous uncertainty estimation}}: For the uni-modal target, we focus on content quality uncertainty; for the multi-modal query, we consider both content quality and multi-modal coordination uncertainty, which are integrated via a \emph{provable} dynamic weighting mechanism.

\item \textbf{\emph{Uncertainty-guided learning objectives}}:
Beyond holistic contrast, we introduce fine-grained contrastive loss, incorporating \emph{component}-, \emph{instance}-, and \emph{modality}-wise negative sampling strategies to enhance learning efficacy.

\item \textbf{\emph{Empirical results}}: 
Benchmark results validate \modelname{}'s superiority to state-of-the-art. 
Model analyses justify key designs. 
Quantitative study highlights the interpretability.
\end{itemize}

\begin{figure*}[t]
    \centering
    \includegraphics[width=\textwidth]{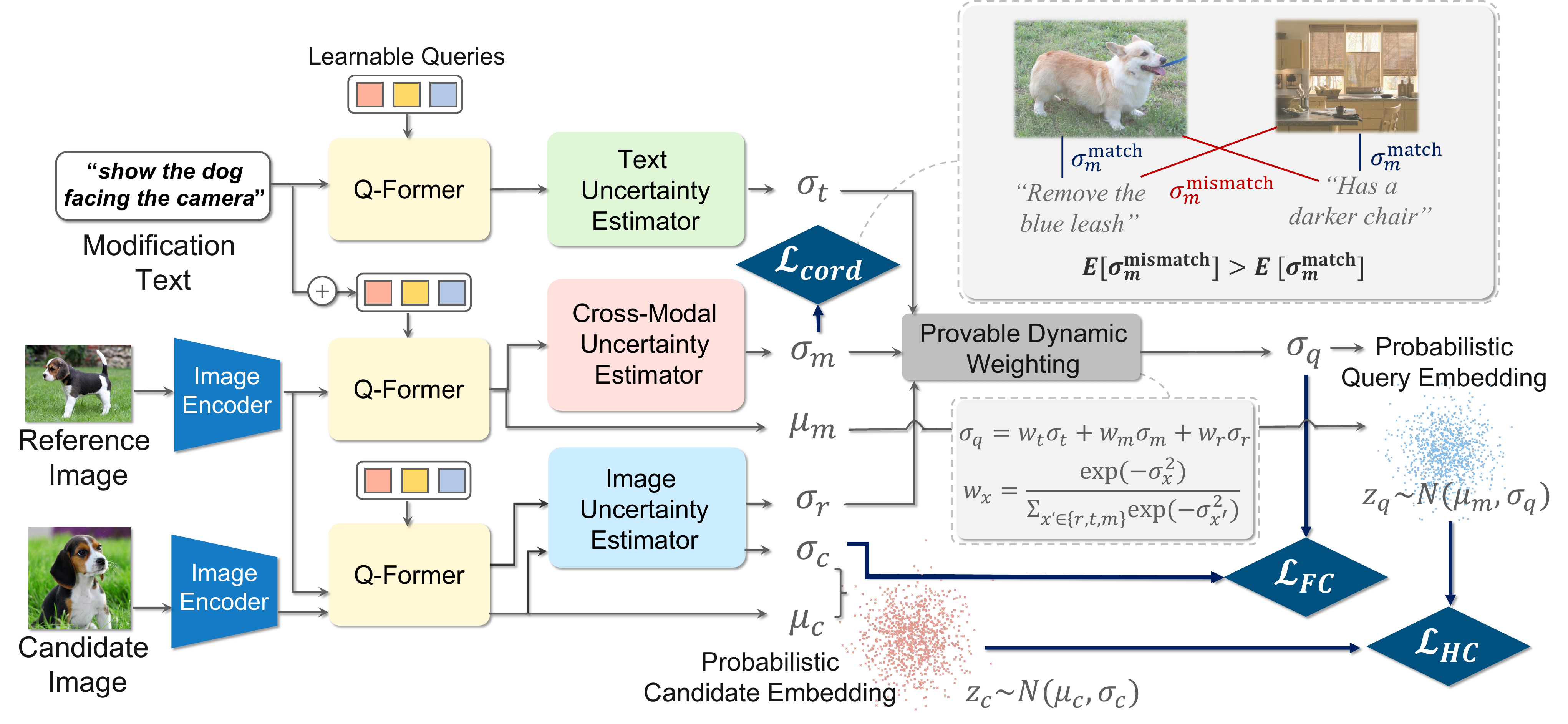}    \caption{\underline{\textbf{\textbf{H}}}eterogeneous \underline{\textbf{\textbf{U}}}ncertainty-\underline{\textbf{\textbf{G}}}uided (\textbf{\modelname{}}) CIR. 
    Modules with the same name share the same weights.}
    \label{fig:arc}
\end{figure*}
\section{Related Works}
\label{sec:related_work}

\subsection{Composed Image Retrieval (CIR)}

CIR has two primary directions. Supervised CIR \cite{ wang2022exploring, zhang2022comprehensive, zhao2022progressive, baldrati2022conditioned, wen2023targetguided, yang2023composed, xu2023multimodal, bai2024sentencelevel} uses triplet training (reference image, modification text, target image) to fuse features and capture visual transformations. Zero-shot CIR \cite{baldrati2023zeroshot, saito2023picword, tang2024contextiw, lin2024finegrained, suo2024knowledgeenhanced, wang2025generative, li2025imagine, tang2025missing, tang2025reasonbefore} trains on independent image-text pairs, converting image features to pseudo-text but lacking triplet supervision, resulting in lower accuracy. We focus on supervised CIR.

Despite triplet supervision benefits, supervised CIR faces data quality issues (noise, ambiguity). Recent solutions include high-quality data refinement \cite{jang2024visual, gu2024compodiff, feng2024improving, ventura2024covr}, semantic decomposition \cite{yang2024decomposing, lin2024finegrained, tian2025ccin}, LLM-based intent clarification \cite{baldrati2023zeroshot, karthik2024visionbylanguage, tian2025ccin, huynh2025collm}, and regularization for ambiguous queries \cite{chen2024composed, xu2024diverse}. Unlike methods eliminating uncertainty, our approach uses probabilistic embeddings to model uncertainties explicitly, incorporating them in training/inference for richer representations and robust training.

\subsection{Uncertainty Learning}

Uncertainty quantifies the likelihood that a model's prediction may be incorrect. 
Two key uncertainty sources exist \cite{kiureghian2009aleatory}: (\textbf{i}) \emph{Epistemic Uncertainty} (reduced by more data/improved architecture) and (\textbf{ii}) \emph{Aleatoric Uncertainty} (from inherent data ambiguity, inevitable even with more data \cite{kendall2017what}). This work focuses on aleatoric uncertainty in CIR, aiming to quantify per-sample uncertainty under fixed data constraints.

To explore aleatoric uncertainty in computer vision, early image classification work \cite{shi2019probabilistic, chang2020data, oh2019modeling} used probabilistic distributions (instead of deterministic points) via lightweight uncertainty heads on pre-trained models, enhancing robustness and accuracy \cite{wang2024robust,fang2025grounding}. Subsequent research \cite{song2019polysemous, chun2021probabilistic, chun2024improved} extended this to cross-modal retrieval. However, these methods use late fusion of independent unimodal predictions \cite{gao2024embracing, chen2024composed}, neglecting modality interaction uncertainty and using coarse-grained instance-level estimation—limiting capture of complex dynamics critical to CIR (e.g., concept modification).

Closely related is \cite{xu2024diverse}, addressing CIR's many-to-many correspondence and sparse annotations via identical uncertainty estimation for queries/targets and Monte Carlo sampling. Our approach differs in: (\textbf{i}) a heterogeneous uncertainty framework for queries-side capturing multi-modal coordination uncertainty; (\textbf{ii}) a closed-form uncertainty-aware distance metric computing expected query-target distance, improving efficiency and stability.
\section{Our Solution}
\label{sec:method}

\subsection{Problem Formulation and Method Overview}
\label{subsec:overview}
Composed Image Retrieval (CIR) operates on triplet data. 
Given a triplet $(x_r, x_t, x_c)$, where \(x_r\) denotes the reference image, \(x_t\) denotes the attached modification text, and \(x_c\) denotes the matched target image. 
The goal of CIR models is to learn a pair of encoders, $f_q$ and $f_c$, producing multi-modal query representation \(z_q = f_q(x_r, x_t)\) and image target representation \(z_c = f_c(x_c)\), such that the query is closer to the target image than to any other candidate images: 
\begin{equation}
    d(z_q, z_c) < d(z_q, z_{c'}), \quad z_c \neq z_{c'}.
\end{equation}
$d(\cdot,\cdot)$ denotes the distance metric. 

Considering various forms of uncertainties caused by data noise in CIR, we propose a \underline{\textbf{H}}eterogeneous \underline{\textbf{U}}ncertainty-\underline{\textbf{G}}uided paradigm (\textbf{\modelname{}}) based on probabilistic learning.
Specifically, we represent each query and target as a series of Gaussian embeddings. 
Take the query $(x_r,x_t)$ as an example, its representation $z_q$ is defined by $[z_q^1,z_q^2,\cdots, z_q^K]$, where the $k$-th sub-representation is parameterized by a Guassian, namely $z_q^k\sim\calN(\mu_q^k,\Sigma_q^k)$, $\mu_q^k\in\bbR^D$, $\Sigma_q^k\in\bbR^{D\times D}$. 
For computation efficiency, we follow common practice \cite{song2019polysemous, chun2021probabilistic, chun2024improved} that simplifies the covariance matrix $\Sigma_q^k$ as a diagonal matrix by assuming dimensional mutual independence, and thus $z_q^k\sim\calN(\mu_q^k,{\sigma_q^k}^2\mathrm{I})$. ${\sigma_q^k}^2\in\bbR^D$ is the variance vector reflecting the uncertainty and $\mathrm{I}$ denotes the identity matrix. 

As shown in \Cref{fig:arc}, we use BLIP-2's Q-Former \cite{li2023blip2} to extract the fine-grained mean vectors via Q-Former's learnable query tokens, where each of the $K=32$ tokens corresponds to a Gaussian. 
On the query side, the visual information is extracted with a pre-trained and fixed visual backbone and injected as the key and value in the cross-attention layers. 
We formulate this process by
\begin{equation}
    \mu_q = h(x_\texttt{[LQ]},x_t,x_r)\in\bbR^{32\times D},
\end{equation}
where $h(\cdot,\cdot,\cdot)$ denotes the shared Q-Former. $x_\texttt{[LQ]}\in\bbR^{32\times D}$ denotes learnable query tokens in Q-Former. 
On the target side, we leave the modification text blank and extract target mean vectors as 
\begin{equation}
    \mu_c = h(x_\texttt{[LQ]},\emptyset,x_c)\in\bbR^{32\times D}.
\end{equation}

In the following sub-sections, we will first present the heterogeneous uncertainty estimation strategies on the query and target sides, after which we will introduce the uncertainty-guided learning framework and the objectives. 

\subsection{Heterogeneous Uncertainty Estimation} \label{sec:heterogeneous}
Unlike common multi-modal retrieval tasks, CIR features an asymmetric matching between multi-modal queries and uni-modal targets. Thus, we estimate the uncertainties in a heterogeneous manner. 

\subsubsection{Target Uncertainty regarding Visual Content Quality}
On the target side, variance parameters $\sigma_c^2$ indicate fine-grained content quality and visual informativeness from various aspects. 
Following~\cite{chun2024improved}, we employ a 1-layer light-weight Transformer block upon Q-Former's output as the uncertainty (\ie variance) estimator,
\begin{equation}
    \sigma^2_c=g_V(\mu_c)\in\bbR^{32\times D}, 
\end{equation}
where $g_V$ denotes the visual uncertainty estimator. 

\subsubsection{Query-Side Uncertainties regarding Uni-modal Quality and Multi-modal Coordination} \label{sec:coordination}
On the query side, we consider more comprehensive uncertainty estimation. 
We regard the combination of the reference image and the modification text as the text-conditioned image representation:
\begin{align}
z_q = f_q(x_r,x_t) = f_{x_t}(x_r),
\end{align}
where we characterize three uncertainties: 
(\textbf{i}) \emph{Uncertainty in the reference image \( x_r \)}: content quality and visual informativeness of the reference image;
(\textbf{ii}) \emph{Uncertainty in the modification text \( x_t \)}: clarity and specificity of the textual modification;
(\textbf{iii}) \emph{Uncertainty in the text-conditioned function \( f_{x_t}(\cdot) \)}: coordination between modification intent and the reference image.
We argue that uncertainty caused by the modifier function \( f_{x_t}(\cdot) \) arises from intrinsic interactions between the reference image and the modification text, which is beyond the naive combination of uni-modal uncertainties. 
Accordingly, we extract the uncertainty factors as follows
\begin{gather}
    \label{equ:sigma_r}
    \sigma^2_r=g_V(h(x_\texttt{[LQ]},\emptyset,x_r))\in\bbR^{32\times D}, \\
    \label{equ:sigma_t}
    \sigma^2_t=g_T(h(x_\texttt{[LQ]},x_t,\emptyset))\in\bbR^{32\times D}, \\
    \label{equ:sigma_m}
    \sigma^2_m:=\sigma^2_m(x_r,x_t)=g_M(\mu_q)\in\bbR^{32\times D}.
\end{gather}
$g_V$ is the visual uncertainty estimator, sharing weights with its target-side counterpart. $g_T$ and $g_M$ are the textual and multi-modal uncertainty estimators, adopting the same model architecture but being independently parameterized.

To shape a more precise estimation of multi-modal coordination uncertainty, we further introduce regularization for the multi-modal uncertainty estimator. 
Intuitively, image-text pairs from the same triplet should exhibit lower coordination uncertainty than those from different triplets. 
Based on this insight, we design a ranking loss that discriminates the estimated uncertainty of image-text pairs within the same triplet from those across different triplets:
\begin{align}\label{eq:Cord.}
\resizebox{.9\hsize}{!}{
    $\mathcal{L}_{\text{Cord.}} = -\bbE_{(x_r,x_t,x_c)\ne(x_r',x_t',x_c')}
    \log\mathcal{S}\Bigl(\bar{\sigma}_{m}^2(x_r, x_t) - \bar{\sigma}_{m}^2(x_r, x_t')\Bigr)$,
}
\end{align}
where \(\mathcal{S}(x) = \frac{1}{1 + e^{-x}}\) is the sigmoid function. \(\bar{\sigma}_{m}^2(x_c, x_t)\) denotes the mean value of multi-modal coordination uncertainty between $x_c$ and $x_t$.
The devised $\mathcal{L}_{\text{Cord.}}$ encourages the multi-modal uncertainty estimator to predict a higher uncertainty when the correspondence between the reference image and modification text is low or ambiguous.

\subsubsection{Summarized Query Uncertainty via Dynamic Weighting}
\label{sec:fusion}
We combine the multi-modal coordination uncertainty $\sigma_m^2$ with the uni-modal uncertainties $\sigma_r^2$ (reference image) and $\sigma_t^2$ (text) in an element-wise manner to establish the overall query uncertainty. 
This combination is conducted on each of the fine-grained uncertainty components in parallel. 
Since the uncertainties in different aspects are expected to be decoupled and mutually independent, we combine them with a linear combination with dynamic weighting, formulated as
\begin{align}\label{eq:query_dw}
\resizebox{.89\hsize}{!}{
    ${\sigma_q^k[i]}^2 = \sum_{x \in \{r,t,m\}} w_x^k[i]\cdot{\sigma_x^k[i]}^2, \, 1\le k\le 32,\, 1\le i\le D$.
}
\end{align}
The fusion weights are input-adaptive:
\begin{align}
    \label{eq:dweight}
    w_x^k[i] = \frac{\exp\bigl(-{\sigma_x^k[i]}^2\bigr)}{\sum_{x'\in\{r,t,m\}} \exp\bigl(-{\sigma_{x'}^k[i]}^2\bigr)}, 
\end{align}
and satisfy: $w_x^k[i] \ge 0$, $\sum_{x \in \{r,t,m\}} w_x^k[i]=1$. 
Inspired by \citet{zhang2023provable}, we can prove that dynamic fusion using \Cref{eq:dweight} yields a tighter generalization error bound than using any static fusion weights.

\newtheorem{prop}{Proposition}
\begin{prop}[Generalization Error Bounds]
Consider a loss function \( \ell \) that is convex \wrt \underline{\textbf{scalar}} variance values \( \sigma_x^2,\,{x\!\in\!\{r,t,m\}} \). Given 
a training set $\calD$ of size \( N \), let
\(\hat{\mathbb{E}}[\ell(\sigma_x^2)] := \frac{1}{N} \sum_{n=1}^{N} \ell(\sigma_x^2(n))\)
be the empirical estimate of the expected generalization loss across all data, then, for any \( \delta \in (0,1) \), with probability at least \( 1 - \delta \), the following generalization error bound holds:
\begin{align}\label{eq:gerror-dynamic}
\mathcal{E} \leq 
\sum_{x \in \{r,t,m\}} & 
\big[\mathbb{E}(w_x)\cdot \hat{\mathbb{E}}[\ell(\sigma_x^2)] + 
\mathbb{E}(w_x)\cdot \mathfrak{R}_x(\ell({\sigma_x^2}))  \notag \\
& + \mathrm{Cov}(w_x, \ell(\sigma_x^2))\big] + 
3 \sqrt{\frac{\ln(1/\delta)}{2N}}.
\end{align}
where \(\mathbb{E}(w_x)\) is the expectation of fusion weights, \(\mathfrak{R}_x(\ell({\sigma_x^2}))\) is the Rademacher complexity, \(\mathrm{Cov}(w_x, \ell(\sigma_x^2))\) is the covariance between fusion weights and loss values.
\end{prop}

\begin{proof}
    See Appendix.
\end{proof}

\begin{corollary}\label{corollary}
If all the following conditions hold:
(\textbf{i}) \(\ell\) is convex \wrt \textbf{\underline{scalar}} $\sigma_x^2$; 
(\textbf{ii}) \(\ell\) penalizes elements with large uncertainty values, \ie $\rho(w_x^{\text{dynamic}}, \ell(\sigma_x^2)) < 0$, where \(\rho\) is the Pearson Correlation Coefficient; 
(\textbf{iii}) the expectation of dynamic weights is the same as the static weights for each modality, \ie $\mathbb{E}(w_x^{\text{dynamic}}) = w_x^{\text{static}}$, 
then, 
dynamic weights fusion as \cref{eq:dweight} will yield a strictly tighter generalization error bound than the static fusion, \ie $\calE_{\text{dynamic}} < \calE_{\text{static}}$.
\end{corollary}

\begin{proof}
    See Appendix.
\end{proof}

\noindent\textbf{Remarks: Implications for \modelname{}.}
We now analyze how the conditions stated in Corollary~\eqref{corollary} are fulfilled by our proposed method. 
First, the sigmoid loss we adopted, which will be introduced as \Cref{eq:qtr}, is convex and ensures the validity of condition (\textbf{i}). Condition (\textbf{ii}) is supported by the probabilistic learning scheme, which—according to the gradient-based analysis in~\cite{chun2021probabilistic}—leads to the down-weighting of items with higher predictive uncertainty during training. 
Lastly, the structure of \Cref{eq:dweight} guarantees the existence of a subset of dynamic weights $w_x^{\text{dynamic}}$ such that the expectation satisfies $\mathbb{E}(w_x^{\text{dynamic}}) = w_x^{\text{static}}$, thereby meeting condition (\textbf{iii}). Taken together, these observations theoretically establish the superiority of dynamic weighting over static weighting.

\begin{table*}[t]
\centering\resizebox{0.95\linewidth}{!}{
\begin{tabular}{cccccccccc}
\toprule
\multirow{2}{*}{\textbf{Method}} & \multicolumn{2}{c}{\textbf{Dress}}      & \multicolumn{2}{c}{\textbf{Shirt}}      & \multicolumn{2}{c}{\textbf{Top \& Tee}}   & \multicolumn{3}{c}{\textbf{Avg.}}                          \\
\cmidrule(lr){2-3} \cmidrule(lr){4-5} \cmidrule(lr){6-7} \cmidrule(lr){8-10}
& R@10  & R@50  & R@10  & R@50  & R@10  & R@50  & R@10  & R@50  & Avg.  \\
\midrule
CLIP4CIR~\cite{baldrati2022conditioned} & 33.81 & 59.40 & 39.99 & 60.45 & 41.41 & 65.37 & 38.40 & 61.74 & 50.07 \\
ComqueryFormer~\cite{li2024multigrained}& 28.85 & 55.38 & 25.64 & 50.22 & 33.61 & 60.48 & 29.37 & 55.36 & 42.36 \\
CRN~\cite{yang2023composed}             & 32.67 & 59.30 & 30.27 & 56.97 & 37.74 & 65.94 & 33.56 & 60.74 & 47.15 \\
FAME-ViL~\cite{han2023famevil}          & 42.19 & 67.38 & 47.64 & 68.79 & 50.69 & 73.07 & 46.84 & 69.75 & 58.29 \\
MANME~\cite{xu2023multimodal}           & 31.26 & 57.66 & 26.37 & 47.94 & 32.33 & 59.31 & 29.99 & 54.97 & 42.48 \\
DWC~\cite{huang2024dynamic}             & 32.67 & 57.96 & 35.53 & 60.11 & 40.13 & 66.09 & 36.11 & 61.39 & 48.75 \\
CompoDiff$^{\bigstar}$~\cite{gu2024compodiff}  & 40.65 & 57.14 & 36.87 & 57.39 & 43.93 & 61.17 & 40.48 & 58.57 & 49.53 \\
MGUR~\cite{chen2024composed}            & 32.61 & 61.34 & 33.23 & 62.55 & 41.40 & 72.51 & 35.75 & 65.47 & 50.61 \\
SSN~\cite{yang2024decomposing}          & 34.36 & 60.78 & 38.13 & 61.83 & 44.26 & 69.05 & 38.92 & 63.89 & 51.40 \\
BLIP4CIR+Bi~\cite{liu2024bidirectional} & 42.09 & 67.33 & 41.76 & 64.28 & 46.61 & 70.32 & 43.49 & 67.31 & 55.40 \\
SPIRIT~\cite{chen2024spirit}            & 39.86 & 64.30 & 44.11 & 65.60 & 47.68 & 71.70 & 43.88 & 67.20 & 55.54 \\
SADN~\cite{wang2024semantic}            & 40.01 & 65.10 & 43.67 & 66.05 & 48.04 & 70.93 & 43.91 & 67.36 & 55.63 \\
CaLa~\cite{jiang2024cala}               & 42.38 & 66.08 & 46.76 & 67.28 & 50.93 & 74.11 & 46.69 & 69.16 & 57.92 \\
CASE$^{\bigstar}$~\cite{levy2024data}    & \textbf{48.48} & 70.23 & 47.44 & 69.36 & 50.18 & 72.24 & 48.70 & 70.61 & 59.66 \\
CoVR$^{\bigstar}$~\cite{ventura2024covr} & -     & -     & -     & -     & -     & -     & 49.40 & 70.98 & 60.19 \\
VDG$^{\bigstar\spadesuit}$~\cite{jang2024visual} & 47.89 & 69.81 & 51.36 & 71.08 & 53.29 & 74.65 & 50.85 & 71.85 & 61.35 \\
QuRe~\cite{kwak2025qure}                & 46.80 & 69.81 & \textbf{53.53} & \underline{72.87} & \underline{57.47} & \underline{77.77} & \underline{52.60} & 
\underline{73.48} & \underline{63.04} \\
\hline
\textbf{\modelname{} (Ours)} & \underline{48.37} & \textbf{71.56} & \underline{51.62} & \textbf{74.41} & \textbf{58.26} & \textbf{78.22} & \textbf{52.75} & \textbf{74.73} & \textbf{63.74}\\
\bottomrule
\end{tabular}}
\caption{Comparison with existing methods on Fashion-IQ dataset. The best results are in bold font and second best results are underlined. Methods using extra data are marked with $^{\bigstar}$ and methods using an LLM with $^{\spadesuit}$.}
\label{tab:comparison}
\end{table*}

\begin{table*}[t]
\label{tab:cirr}
\centering\resizebox{0.9\linewidth}{!}{
\begin{tabular}{ccccccccc}
\toprule
\multirow{2}{*}{\textbf{Method}} & \multicolumn{4}{c}{\textbf{Recall@K}} & \multicolumn{3}{c}{\(\textbf{Recall}_{subset}\)\textbf{@K}} & \multirow{2}{*}{\(\frac{\textbf{R@5}+\textbf{R}_s\textbf{@1}}{2}\)} \\ 
\cmidrule(lr){2-5} \cmidrule(lr){6-8}
& K=1 & K=5 & K=10 & K=50 & K=1 & K=2 & K=3 & \\ 
\midrule
CIRPLANT~\cite{liu2021image} & 19.55 & 52.55 & 68.39 & 92.38 & 39.20 & 63.03 & 79.49 & 45.88 \\
CompoDiff$^{\bigstar}$~\cite{gu2024compodiff} & 32.39 & 57.61 & 77.25 & 94.61 & 67.88 & 85.29 & 94.07 & 62.75 \\
CASE$^{\bigstar}$~\cite{levy2024data} & 49.35 & 80.02 & 88.75 & 97.47 & 76.48 & 90.37 & 95.71 & 78.25 \\
VDG$^{\bigstar\spadesuit}$~\cite{jang2024visual} & 50.96 & 80.15 & 86.86 & 94.46 & 77.45 & 90.65 & 96.10 & 78.80 \\
ComqueryFormer~\cite{li2024multigrained} & 25.76 & 61.76 & 75.90 & 95.13 & 51.86 & 76.26 & 89.25 & 56.81 \\
CLIP4CIR~\cite{baldrati2022conditioned} & 38.53 & 69.98 & 81.86 & 95.93 & 68.19 & 85.64 & 94.17 & 69.09 \\
MANME~\cite{xu2023multimodal} & 18.27 & 48.02 & 63.23 & 89.66 & 42.43 & 64.89 & 77.93 & 45.23 \\
SPIRIT~\cite{chen2024spirit} & 40.32 & 75.10 & 84.16 & 96.88 & 73.74 & 89.60 & \underline{95.93} & 74.42 \\
SSN~\cite{yang2024decomposing} & 43.91 & 77.25 & 86.48 & 97.45 & 71.76 & 88.63 & 95.38 & 74.51 \\
SADN~\cite{wang2024semantic} & 44.27 & 78.10 & 87.71 & 97.89 & 72.71 & 89.33 & 95.38 & 75.41 \\
QuRe~\cite{kwak2025qure} & \textbf{52.22} & \underline{82.53} & \underline{90.31} & \textbf{98.17} & \underline{78.51} & \underline{91.28} & \textbf{96.48} & \underline{80.52} \\
\hline
\textbf{\modelname{} (Ours)} & \underline{51.09} & \textbf{83.20} & \textbf{92.03} & \underline{97.89} & \textbf{80.65} & \textbf{91.80} & \underline{95.93} & \textbf{81.93} \\
\bottomrule
\end{tabular}}
\caption{Comparison with existing methods on CIRR dataset. The best results are in bold font and second best results are underlined. Methods using extra data are marked with $^{\bigstar}$ and methods using an LLM with $^{\spadesuit}$.}
\label{tab:cirr}
\end{table*}

\subsection{Uncertainty-Guided Learning}\label{sec:learning}
\subsubsection{Uncertainty-Guided Holistic Query-Target Contrast}
We adopt a sigmoid contrastive loss \cite{zhai2023sigmoid} to holistically align the query and target representations.
\begin{align}\label{eq:qtr}
\mathcal{L}_{\text{HC}} = & - \bbE_{(x_r,x_t,x_c)}\log\big(\mathcal{S}(-a\!\cdot\! d(z_q, \!z_c)\! - \! b)\big)\big) \notag \\
& - B\cdot\bbE_{(x_r',x_t',x_c')\ne(x_r,x_t,x_c)}\log\big(\mathcal{S}(a\!\cdot\! d(z_q, \!z_c')\! + \! b)\big)\big) \notag \\
& - B\cdot\bbE_{(x_r',x_t',x_c')\ne(x_r,x_t,x_c)}\log\big(\mathcal{S}(a\!\cdot\! d(z_q',\! z_c)\! + \! b)\big)\big), 
\end{align}
where $\calS(\cdot)$ is the Sigmoid function. $B$ represents the proportion of negative samples to positive samples, typically set as the batch size. 
$d(\cdot, \cdot)$ denotes the uncertainty-aware holistic distance metric between queries and target images. 
$a$ and $b$ are two learnable parameters initialized by $1$ and $0$.
Following related works \cite{shi2019probabilistic, chang2020data, oh2019modeling}, we compute the uncertainty-aware distance as the expected Euclidean distance between two Gaussians. 
Consider two points \(z_1\sim\mathcal{N}(\mu_1, \sigma_1^2\mathrm{I})\) and \(z_2\sim\mathcal{N}(\mu_2, \sigma_2^2\mathrm{I})\), their expected Euclidean distance is:
\begin{align}
\resizebox{.89\hsize}{!}{
    $\mathbb{E}_{z_1, z_2} \big[||z_1 - z_2||_2^2\big] = ||\mu_1 - \mu_2||_2^2 + ||\sigma_1||_2^2 + ||\sigma_2||_2^2$.
}
\end{align}
By applying the above distance metric to each fine-grained component of the query and target embeddings, we can derive the uncertainty-aware holistic distance metric as
\begin{align}
    d(z_q, z_c) = ||\mu_q - \mu_c||_\mathsf{F}^2 + ||\sigma_q||_\mathsf{F}^2 + ||\sigma_c||_\mathsf{F}^2.
\end{align}
Here, $\|\cdot\|_\mathsf{F}$ denotes the Frobenius norm of the tensor.

\subsubsection{Uncertainty-Guided Fine-Grained Contrast}
In order to align the fine-grained representations between the query and target, and promote the orthogonality and diversity of fine-grained uncertainty components, we introduce a contrastive strategy. 
Specifically, 
for the variance vector of the $k$-th fine-grained component, \(\sigma^k_M \), the loss encourages differentiation:
\begin{align}\label{eq:lfc}
\resizebox{.89\hsize}{!}{
$\mathcal{L}_{\text{FC}}=-\sum_{M\in\{q, c\}}
\sum_{k=1}^{32} 
\bbE_{\sigma^{k'}_{M'}\ne\sigma^k_M}\zhongk{\log\Bigg(\calS\Big( a' \big\| \sigma^k_{M} - \sigma^{k'}_{M'} \big\|_2^2 + b' \Big)\Bigg)}$,
}
\end{align}
where \(a'\) and \(b'\) are learnable. We employ three negative sampling strategies for $\sigma_{M'}^{k'}$:
\textbf{(i)} \emph{Component-wise}: Negatives are other components of the same side and instance.
\textbf{(ii)} \emph{Instance-wise}: Negatives are other components of the same side but different instances.
\textbf{(iii)} \emph{Modality-wise}: Negatives from other components of the other side and any instances.

\subsubsection{Overall Learning Objectives}
Total learning objectives: 
\begin{align}\label{equ:total_objectives}
    \mathcal{L}_\text{\modelname{}} = \mathcal{L}_{\text{HC}} + \lambda_{\text{FC}} \mathcal{L}_{\text{FC}} + 
    \lambda_{\text{Cord.}} \mathcal{L}_{\text{Cord.}}.
\end{align}
$\lambda_{\text{FC}}$ and $\lambda_{\text{Cord.}}$ are loss balancing factors. 
\section{Experiments}
\label{sec:experiments}

\subsection{Research Questions}
\label{subsec:research_questions}
We aim to answer the following research questions by conducting experiments on two standard CIR benchmarks:
\begin{itemize}[leftmargin=3em]
	\item[\textbf{RQ1:}] Compared to state-of-the-art approaches, how does \modelname{} perform on CIR benchmarks?
	\item[\textbf{RQ2:}] How does each design contribute in \modelname{}? 
	\item[\textbf{RQ3:}] How to interpret the efficacy of our proposed \modelname{}?
\end{itemize}

\begin{table*}[t]
    \centering
    \renewcommand{\arraystretch}{1.2}
    \setlength{\tabcolsep}{6pt}
    \resizebox{1.0\linewidth}{!}{
    \begin{tabular}{@{}clccccccccccc@{}}
        \toprule
        \multirow{2}{*}{\#} & \multirow{2}{*}{\textbf{Experimental Variant}} & \multicolumn{2}{c}{\textbf{Dress}} & \multicolumn{2}{c}{\textbf{Shirt}} & \multicolumn{2}{c}{\textbf{Top \& Tee}} & \multicolumn{2}{c}{\textbf{Average}} & \multirow{2}{*}{\textbf{Avg.}} & \multirow{2}{*}{\makecell{\textbf{ Avg. Time} \\ \textbf{/ Query (ms)}}} \\
        \cmidrule(lr){3-4} \cmidrule(lr){5-6} \cmidrule(lr){7-8} \cmidrule(lr){9-10}
        & & R@10 & R@50 & R@10 & R@50 & R@10 & R@50 & R@10 & R@50 & & \\
        \midrule
        \multicolumn{12}{l}{\emph{\textbf{Baselines}}} \\
        (0) & Point Matching & 40.52 & 62.25 & 39.89 & 62.77 & 43.03 & 65.12 & 41.15 & 63.38 & 52.26 & 7.51 \\
        (1) & + Probabilistic Embedding & 42.74 & 64.40 & 44.74 & 65.71 & 47.52 & 67.55 & 45.00 & 65.89 & 55.44 & 10.08\\
        \midrule
        \multicolumn{12}{l}{\emph{\textbf{Fine-Grained Uncertainty-Guided Learning}}} \\
        (2) & + Component-Wise Fine-Grained Contrast & 44.28 & 64.86 & 48.89 & 67.70 & 51.61 & 70.45 & 48.26 & 67.67 & 57.97 & 20.69\\
        (3) & + Instance-Wise Fine-Grained Contrast & 44.64 & 65.47 & 49.73 & 67.85 & 52.03 & 71.15 & 48.80 & 68.16 & 58.48 & 20.54\\
        (4) & + Modality-Wise Fine-Grained Contrast & 45.13 & 66.83 & 49.27 & 68.97 & 53.85 & 71.92 & 49.42 & 69.24 & 59.33 & 20.73\\
        \midrule
        \multicolumn{12}{l}{\emph{\textbf{Heterogeneous Uncertainty Estimation}}} \\
        (5) & + Cross-Modal Uncertainty & 44.15 & 65.68 & 49.04 & 68.28 & 52.96 & 71.31 & 48.72 & 68.42 & 58.57 & 21.28\\
        (6) & + Multi-Modal Coordination Loss & 47.82 & 70.28 & 51.27 & 73.96 & 57.69 & 77.62 & 52.26 & 73.95 & 63.11 & 21.19\\
        (7) &  \textbf{+ Dynamic Weighting (Full Model)} & \textbf{48.37} & \textbf{71.56} & \textbf{51.62} & \textbf{74.41} & \textbf{58.26} & \textbf{78.22} & \textbf{52.75} & \textbf{74.73} & \textbf{63.74} & 21.35\\
        \bottomrule
    \end{tabular}}
    \caption{Ablation Study on the Fashion-IQ dataset. Variants in the table add components progressively from top to bottom. We conduct validation on a single A100 GPU and report the average retrieval time per query (inference + distance computation).}
    \label{tab:ablation}
\end{table*}

\begin{figure}[t]
    \centering
    \includegraphics[width=\columnwidth]{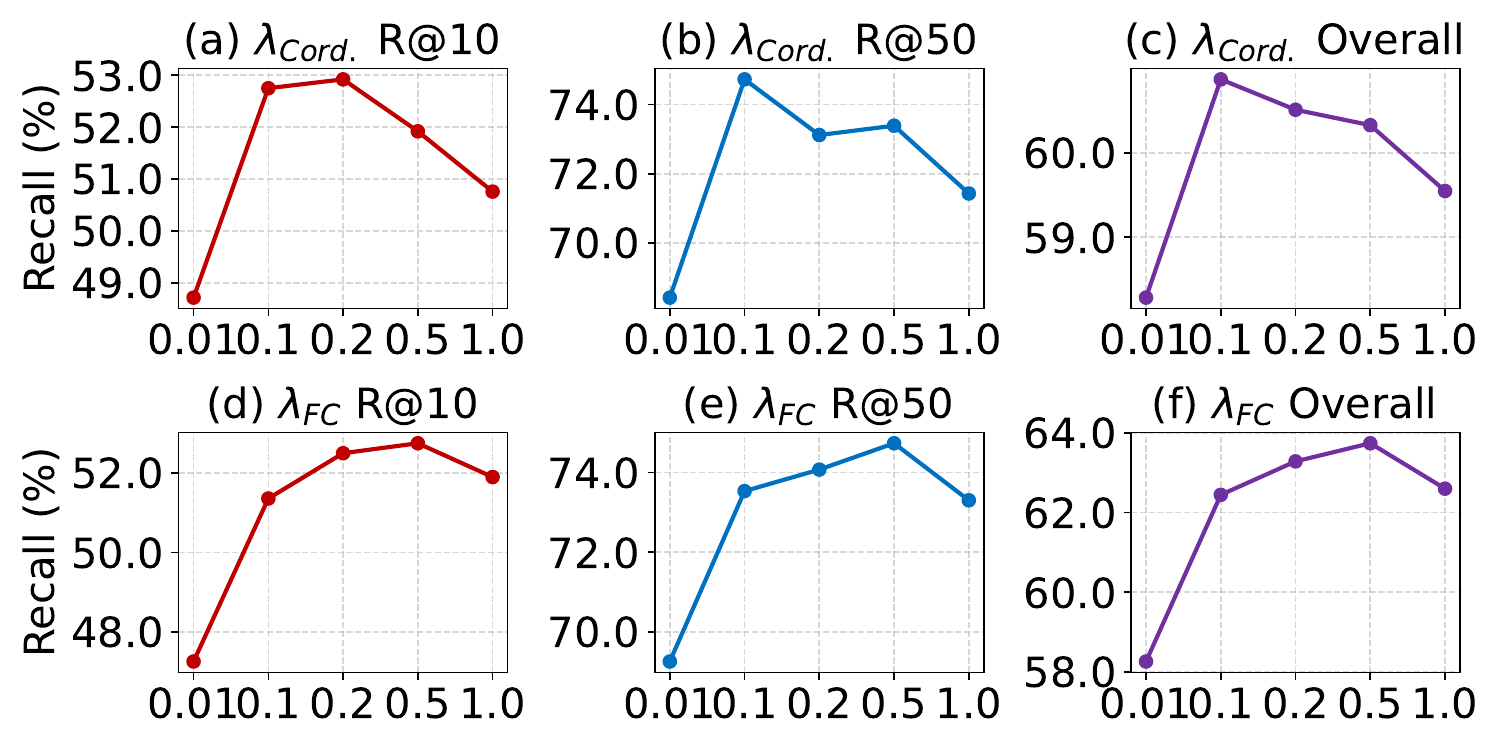 }
    \caption{Model performance (average recall) on Fashion-IQ dataset under different settings of $\lambda_{\mathrm{Cord.}}$ and $\lambda_{\mathrm{FC}}$.}
    \label{fig:hp}
\end{figure}

\begin{figure*}[t]
\centering
\begin{minipage}{0.33\textwidth}
    \centering
    \includegraphics[width=\textwidth]{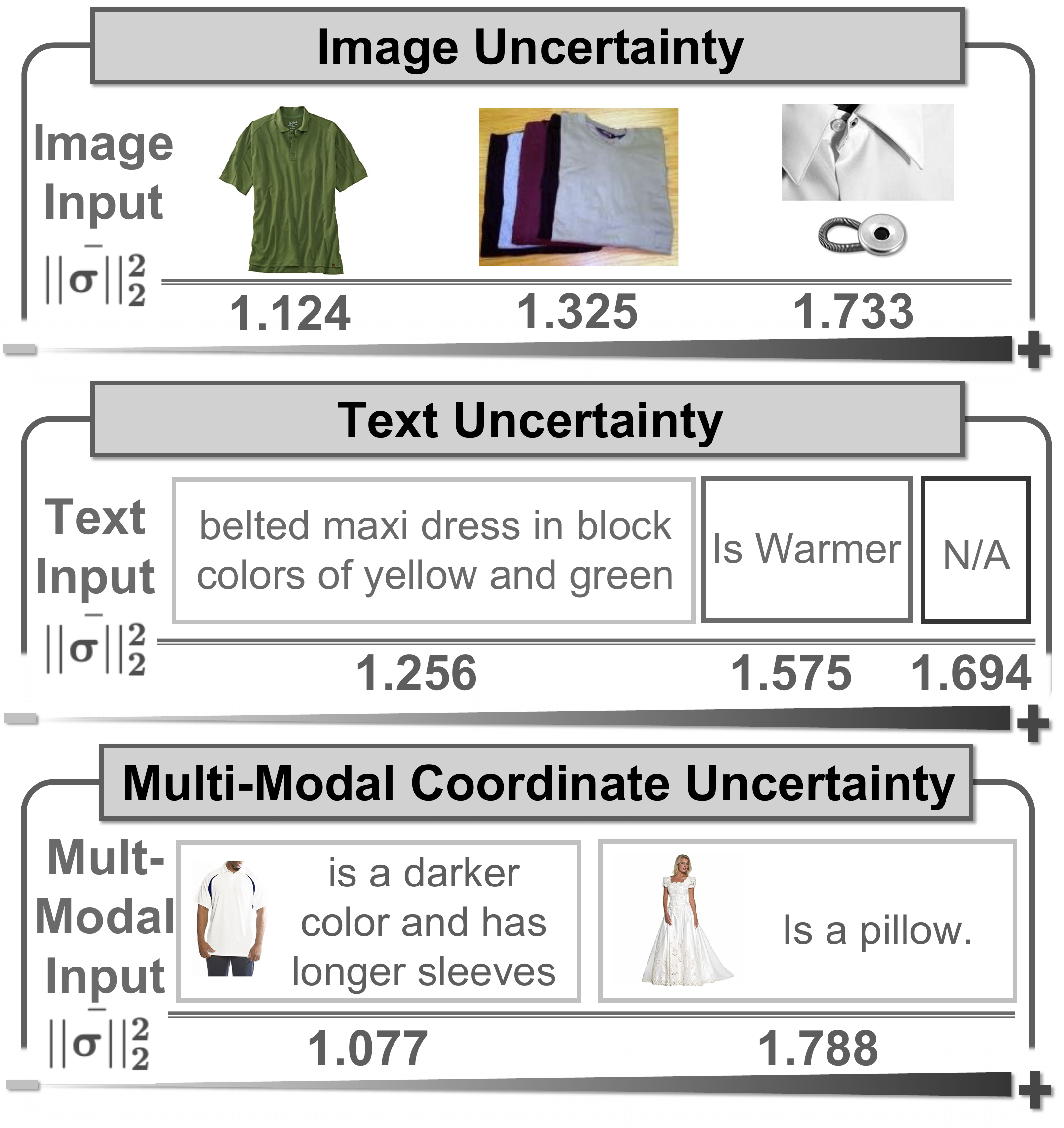}
\end{minipage}
\hspace{.5em}
\begin{minipage}{0.65\textwidth}
    \centering
    \includegraphics[width=\textwidth]{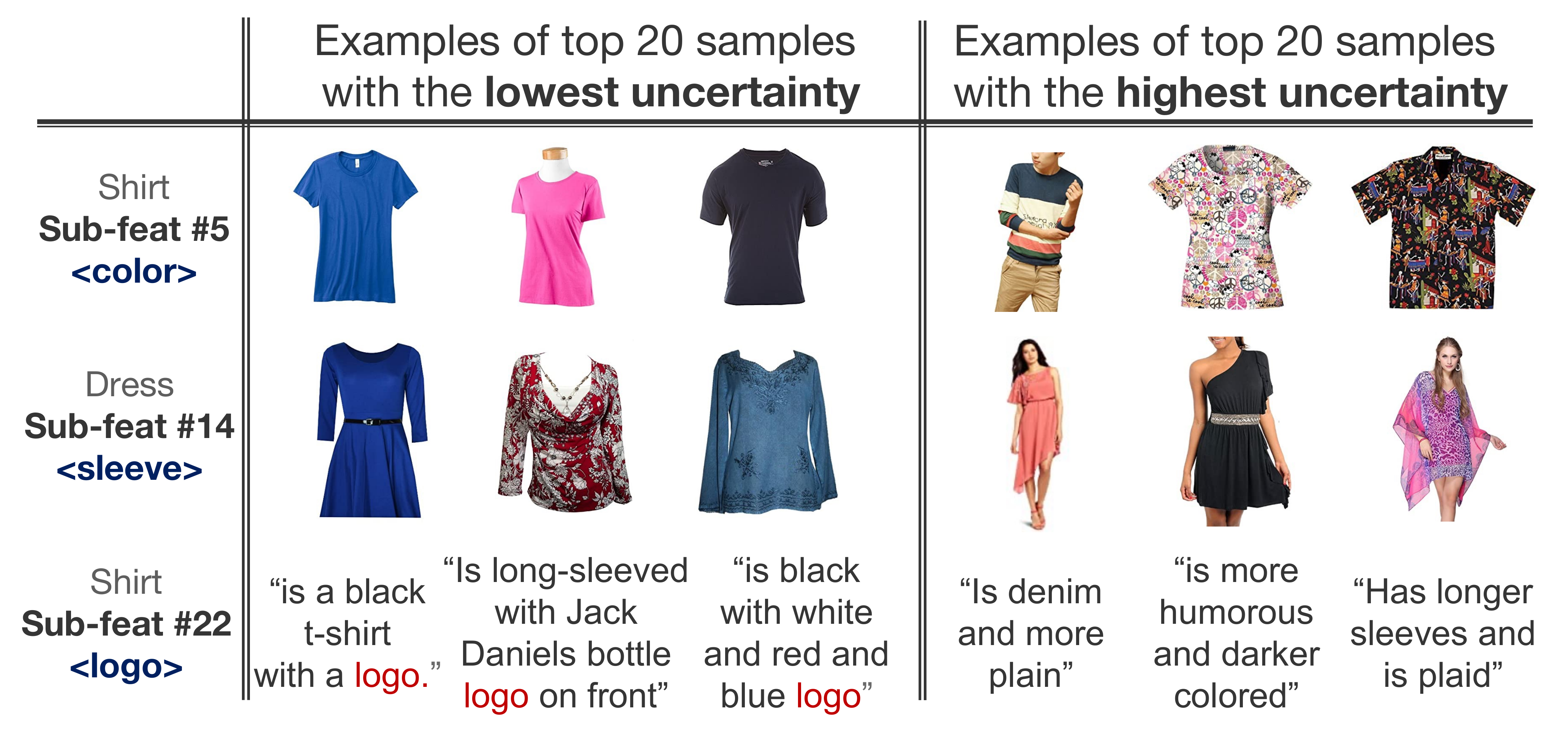}
\end{minipage}
\caption{Qualitative analysis illustrating the meaning behind our learned uncertainty: (Left) Overall level of uncertainty reflects data quality. (Right) Different fine-grained uncertainty component corresponds to different sub-concepts.}
\label{fig:case_study}
\end{figure*}

\subsection{Experimental Setup}
\label{subsec:setup}
\subsubsection{Datasets and Metrics}
\label{subsubsec:dataset_setup}
We evaluate our model on two major benchmarks for composed image retrieval. Fashion-IQ \cite{wu2021fashion} is a fashion dataset consisting of 18,000 training triplets and 6,016 validation triplets, with a total of 15,536 candidate images for validation. Model performance on this dataset is reported using the Recall@K metric for K=10 and K=50. CIRR \cite{liu2021image} comprises 36,554 image triplets derived from 21,552 real-world photographs originally sourced from NLVR2. In addition to the conventional Recall@K metric, CIRR introduces a novel evaluation framework, \(\text{Recall}_{subset}@\text{K}\), which assesses a model’s fine-grained ability to distinguish target images within small groups of six visually similar images. 

\subsubsection{Implementation Details}
\label{subsubsec:baselines_setup}
We employ the pre-trained weights of BLIP-2 as the initial weights of the Q-Former. 
Training is conducted on a single A100-80G GPU with a batch size of 32 and an initial learning rate of $3\times10^{-5}$. 
We implement an AdamW optimizer with parameters $\beta_1=0.9$, $\beta_2=0.999$, $\epsilon=1.0\times 10^{-7}$. 
Default hyper-parameter settings are 
$\lambda_\text{Cord.} = 0.1$ and 
$\lambda_\text{FC} = 0.5$ 
for \cref{equ:total_objectives}.

\subsection{Comparasion with State-of-The-Arts (RQ1)}
\label{subsec:sota}
We conduct comprehensive comparisons against state-of-the-art methods on both Fashion-IQ and CIRR datasets. 
As is shown in tables \ref{tab:comparison} and \ref{tab:cirr}, \modelname{} achieves significant improvement against existing SoTAs across both benchmarks, demonstrating the effectiveness of our proposed uncertainty-guided framework for composed image retrieval tasks. It is worth highlighting that \modelname{} has outperformed methods that applies out-sourced data (e.g., videos, web images, AI generated images), as well as methods that utilize LLMs to refine or rewrite prompts. This indicates that under proper uncertainty-aware supervision, a CIR model can effectively identify noise within training data and achieve robust matching without the need for additional curated labels or LLM-based enhancements.

\subsection{Model Analyses (RQ2)}
\label{subsec:model_analyses}

\subsubsection{Ablation Study} 

We compare configurations against a point matching baseline and a probabilistic embedding baseline. Baseline (0) aligns query image-text embeddings with target image embeddings using InfoNCE loss \cite{he2020momentum}. Probabilistic baseline (1) uses equation~\eqref{eq:qtr} with generalized pooling (GPO) \cite{chen2021learning} for global uncertainty.
Comparing (0) and (1) shows benefits of probabilistic uncertainty modeling.

Experiments (2,3,4) add fine-grained uncertainty-guided learning (equation~\eqref{eq:lfc}) using mean-pooled unimodal uncertainties. Consistent improvements over (1) confirm that finer uncertainty granularity enhances performance.
The contrastive loss components are also verified.

Experiments (5,6,7) investigate heterogeneous uncertainty by integrating cross-modal with unimodal uncertainties. (4,5,6) show naive cross-modal inclusion degrades performance, while our multi-modal coordination loss is critical for improvement—highlighting the need to disentangle cross-modal and unimodal uncertainties. (6) vs (7) shows dynamic weighting outperforms static averaging in fusion.

\subsubsection{Hyper-parameter Sensitivity}  
We analyze the impact of two key hyper-parameters in our learning objective: 
\begin{itemize}
    \item \emph{Coefficient of multi-modal coordination loss, $\lambda_{\mathrm{Cord.}}$.} 
    This coefficient balances the query-target ranking loss and the multi-modal coordination loss. As shown in Figures~\ref{fig:hp}(a-c), setting $\lambda_{\mathrm{Cord.}}$ to 0.1 allows it to act as an effective regularizer. However, increasing its weight too much leads to performance degradation.  
    \item \emph{Coefficient of fine-grained contrastive loss, $\lambda_{\mathrm{FC}}$.}
    This coefficient controls the balance between the query-target ranking loss and the fine-grained uncertainty-contrastive loss. Figures~\ref{fig:hp}(d-f) demonstrate that reducing $\lambda_{\mathrm{FC}}$ results in significant performance degradation, highlighting the importance of fine-grained contrastive loss in capturing meaningful fine-grained uncertainties.
\end{itemize}

\subsection{Interpretability of \modelname{} (RQ3)}
\label{subsec:model_interpretability}

\subsubsection{Understanding overall uncertainty values.} 
 We analyze sample quality across uncertainty levels on Fashion-IQ. Overall uncertainty is defined as \(||\bar{\sigma}||_2^2=\frac{1}{K}\sum||\sigma^k||^2_2\) (average variance of fine-grained components). As shown in Figure~\ref{fig:case_study} (left), higher overall uncertainty correlates with lower sample quality. Multi-modal coordinate uncertainty also reflects image-text correspondence ambiguity, confirming the model assesses both unimodal content quality and multi-modal interaction clarity.

\subsubsection{Understanding fine-grained uncertainty values.}
We study sub-feature uncertainties via case studies: curating top/bottom 20 instances for each sub-feature (filtering high overall uncertainty samples).  
Qualitative analysis (Figure~\ref{fig:case_study}, right) reveals clear links between fine-grained uncertainties and real-world concepts. For example: Fashion-IQ Shirt's 5th sub-feature connects to color; Similar phenomena also occur in Dress's 14th sub-feature and Shirt's 22nd sub-feature. This confirms the model effectively captures fine-grained concept uncertainty.
\section{Conclusions}
\label{sec:conclusion}

We propose a novel \textbf{H}eterogeneous \textbf{U}ncertainty-\textbf{G}uided (\modelname{}) paradigm for Composed Image Retrieval (CIR).
\modelname{} represents both queries and targets as fine-grained Gaussian distributions, where the variances encode heterogeneous uncertainties.
We apply a dynamic weighting mechanism that integrates uncertainty cues from content quality and cross-modal coordination, and formulate effective learning objectives for robust holistic and fine-grained matching.
Extensive experiments demonstrate that \modelname{} consistently outperforms prior approaches, offering resilience to noisy inputs.
Our results highlight the critical role of uncertainty modeling in CIR, providing valuable insights for user-centric visual search systems and offering broader impact for related tasks like 
universal retrieval~\cite{wei2023uniir}. 

\clearpage
\section{Acknowledgments}
We sincerely thank the anonymous reviewers and chairs for their efforts and constructive suggestions, which have greatly helped us improve the manuscript. 
This work is supported in part by the National Natural Science Foundation of China under grants 624B2088 and 62571298. 
Long Chen was supported by the Hong Kong SAR RGC Early Career Scheme (26208924), and the National Natural Science Foundation of China Young Scholar Fund (62402408).

\bibliography{aaai2026}
\clearpage

\section{Proof}

\subsection{Proof of Proposition 1.}

\begin{proof}
Given the convexity of the loss function \( \ell \) \wrt each uncertainty component \( \sigma_x^2 \), 
by Jensen’s inequality, we have:
\begin{align}
    \ell\left(\sum_{x \in \{r,t,m\}} w_x \sigma_x^2 \right) 
    \leq \sum_{x \in \{r,t,m\}} w_x \ell(\sigma_x^2).
\end{align}
By taking the expectation on both sides, we obtain:
\begin{align}\label{eq:jensen_expectation}
   \mathcal{E}\! :=\! \mathbb{E}\!\! \left[\ell\!\!\left(\sum_{x \in \{r,t,m\}} w_x \sigma_x^2\! \right)\!\! \right] 
    \!\!\leq\!\! \sum_{x \in \{r,t,m\}} \mathbb{E} [w_x\ell(\sigma_x^2)].
\end{align}
Note the right part of Eq. (\ref{eq:jensen_expectation}) can be expanded as 
\begin{align}
    \label{equ:covariance_property}
    \mathbb{E} [w_x \ell(\sigma_x^2)] 
    = \mathbb{E}(w_x) \cdot \mathbb{E} [\ell(\sigma_x^2)] 
    + \textrm{Cov}(w_x, \ell(\sigma_x^2)).
\end{align}

According to Rademacher complexity theory~\cite{bartlett2002rademacher}, for \( \delta \in (0,1) \), with probability at least \( 1- \delta \), we have
\begin{align}
    \label{equ:Rademacher}
    \mathbb{E} [\ell(\sigma_x^2)] 
    \leq \hat{\mathbb{E}}[\ell(\sigma_x^2)] 
    + \mathfrak{R}_x(\ell({\sigma_x^2})) 
    + \sqrt{\frac{\ln(1/\delta)}{2N}}.
\end{align}
By combining \cref{equ:Rademacher,equ:covariance_property,eq:jensen_expectation}, we obtain 
\begin{align}\label{eq:gerror-intermediate}
\mathcal{E} \leq 
\sum_{x \in \{r,t,m\}} & 
\big[\mathbb{E}(w_x)\cdot \hat{\mathbb{E}}(\ell(\sigma_x^2)) + 
\mathbb{E}(w_x)\cdot \mathfrak{R}_x(\ell({\sigma_x^2})) \notag \\ 
+ & \mathbb{E}(w_x)\!\cdot\! \sqrt{\frac{\ln(1/\delta)}{2N}} + \mathrm{Cov}(w_x, \ell(\sigma_x^2))\big].
\end{align}
Since $0\leq w_x \leq 1$, $\sum_x\mathbb{E}(w_x)\cdot \sqrt{\frac{\ln(1/\delta)}{2N}} \leq 3 \sqrt{\frac{\ln(1/\delta)}{2N}}$.
Finally, we obtain eq. (12).
\end{proof}

\subsection{Proof of Corollary 1.}

\begin{proof}
For static fusion, constant fusion weights lead to $\mathrm{Cov}(w^\text{static}_x,\ell(\sigma_x^2))=0$, and eq. (12) can be simplified by
\begin{align}\label{eq:gerror-static}
\mathcal{E}_{\text{static}} \leq 
\sum_{x \in \{r,t,m\}} & 
\big[w_x^{\text{static}}\cdot \hat{\mathbb{E}}(\ell(\sigma_x^2)) + 
w_x^{\text{static}}\cdot \mathfrak{R}_x(\ell({\sigma_x^2}))  \big]\notag \\ 
& + 3 \sqrt{\frac{\ln(1/\delta)}{2N}}.
\end{align}

For dynamic fusion, the generalization error bound is
\begin{align}\label{eq:gerror-dynamic}
\mathcal{E}_{\text{dynamic}} \leq 
\sum_{x \in \{r,t,m\}} & 
\big[\mathbb{E}(w_x^{\text{dynamic}})\cdot \hat{\mathbb{E}}(\ell(\sigma_x^2))  \notag \\
& +\mathbb{E}(w_x^{\text{dynamic}})\cdot \mathfrak{R}_x(\ell({\sigma_x^2}))  \notag \\
& + \mathrm{Cov}(w_x^{\text{dynamic}}, \ell(\sigma_x^2))\big] \notag \\
+ & 3 \sqrt{\frac{\ln(1/\delta)}{2N}}.
\end{align}
By comparing \cref{eq:gerror-dynamic,eq:gerror-static}, and given that \(\mathbb{E}(w_x^{\text{dynamic}}) = w_x^{\text{static}}\), we can infer that 
\begin{align}
    \sup \mathcal{E}_{\text{dynamic}} \leq \sup \mathcal{E}_{\text{static}},
\end{align}
under the condition 
\begin{align}
    \sum_{x\in\{r,t,m\}} \mathrm{Cov}\!\big(w_x^{\text{dynamic}},\,\ell(\sigma_x^2)\big)\;<\;0.
\end{align}

This condition is guaranteed because eq. (11) imposes \textbf{reduced} modality weight when the modality uncertainty increases, while the modality loss component \(\ell(\sigma_x^2)\) will \textbf{increases} when uncertainty increase. Subsequently, we can establishing that the dynamic fusion achieves a generalization error bound that is no worse (and ideally better) than any static fusion schemes.
\end{proof}

\end{document}